\documentclass[letterpaper]{article} 
\usepackage{aaai2027}  
\usepackage[hyphens]{url}  
\usepackage{graphicx} 
\usepackage{natbib}  
\usepackage{caption} 
\usepackage{algorithm}
\usepackage{algorithmic}

\usepackage{newfloat}
\usepackage{listings}
\DeclareCaptionStyle{ruled}{labelfont=normalfont,labelsep=colon,strut=off} 
\floatstyle{ruled}
\newfloat{listing}{tb}{lst}{}
\floatname{listing}{Listing}

\usepackage{booktabs}
\usepackage{multirow}
\usepackage{array}
\usepackage{amsmath}
\usepackage{amssymb}

\title{CCDWIKI: Decoupling Case Formulation\\
from Response Generation in Longitudinal CBT Dialogue}
\author{
    Mingwen Zhang \textsuperscript{\rm 1},
    Ruiyang Song \textsuperscript{\rm 1}, 
    Qinying Wang \textsuperscript{\rm 1}, 
    Yuchen Wang \textsuperscript{\rm 1}, 
    Changsheng Ma \textsuperscript{\rm 1, *},
    Minqiang Yang \textsuperscript{\rm 1, *},
}
\affiliations{
    \textsuperscript{\rm 1}School of Information Science and Engineering, Lanzhou University, Lanzhou, China\\

    \{zhangmw2024\}@lzu.edu.cn
}

\begin{document}

\maketitle

\begin{abstract}
Cognitive Behavioral Therapy (CBT) demands more than empathic validation, requiring instead a structured formulation of the interplay among situations, automatic thoughts, emotions, and behaviors. Therapists formalize this interplay through a Cognitive Conceptualization Diagram, where maintaining longitudinal coherence necessitates continuous evidential adjudication to separate tentative hypotheses from established findings. Current dialogue systems track conversational history via retrieval or static profiles but lack the representational capacity to encode epistemic status. Consequently, provisional interpretations are often prematurely reified as established facts, inadvertently biasing downstream interventions. Addressing this limitation, CCDWiki is introduced as a plug-and-play advisor conceptualizing longitudinal therapy as evidence-grounded hypothesis adjudication. The approach encodes each clinical hypothesis as a mutable entry bound to corroborating and contradicting evidence. Across sessions, accumulating data strengthens supported entries, preserves provisional ones, and retracts contradicted assumptions. The resulting formulation distills into concise guidance for unmodified therapeutic language models. Experiments demonstrate substantial improvements in both conceptualization and intervention quality without requiring backbone retraining.
\end{abstract}

\section{Introduction}
\label{sec:introduction}

Cognitive Behavioral Therapy (CBT) is fundamentally distinct from open-ended emotional support. While general supportive conversation aims to provide empathy and comfort, CBT is a highly structured clinical protocol designed to help patients identify, examine, and restructure the maladaptive thoughts and core beliefs that drive their psychological distress \citep{beck2020cognitive,hofmann2012efficacy}. The therapeutic goal is not merely to make the patient feel heard in the immediate turn, but to systematically map their cognitive patterns and guide them toward lasting cognitive restructuring.

Patients rarely arrive at therapy with a clear understanding of their own cognitive architecture; they typically present only surface-level distress and fragmented events. Consider a patient who reports, `I feel anxious because my partner replied late.'' A standard empathetic LLM might respond, `That sounds stressful, and it is understandable.'' This acknowledges the emotion but remains at the surface level. A CBT therapist must instead initiate structured inquiry to uncover the underlying \emph{automatic thought}. If the thought is `I am unlovable,'' the intervention targets core beliefs about self-worth. If it is `Something terrible must have happened to him,'' the intervention shifts toward behavioral testing of catastrophic predictions. To manage this complexity, therapists rely on a Cognitive Conceptualization Diagram (CCD), which provides a structured and evolving representation that links triggering situations, automatic thoughts, emotional responses, and behaviors into a causal chain that directly informs the next clinical intervention \citep{persons1989cognitive,bieling2003case,padesky2020collaborative}.

Building and maintaining an accurate CCD across dialogue turns is a continuous, dynamic process of \emph{evidential adjudication}. A patient does not provide the complete CCD at once; instead, they reveal clinical fragments over multiple turns or sessions. The therapist must continually evaluate these fragments against the current working hypothesis. Does this new statement support the tentative assumption about the patient's core belief, or does it contradict it? Should the hypothesis be promoted to an established clinical fact that guides the next intervention, or should it be revised or retracted? This ongoing judgment involves tracking which evidence supports each tentative hypothesis and determining when that hypothesis should be promoted, revised, or withdrawn. It constitutes the central cognitive task of CBT case formulation. Even experienced human therapists find it highly challenging to maintain this evidential trail accurately \citep{bieling2003case,persons2008caseformulation}.

\begin{figure}[t]
\centering
\includegraphics[width=\columnwidth]{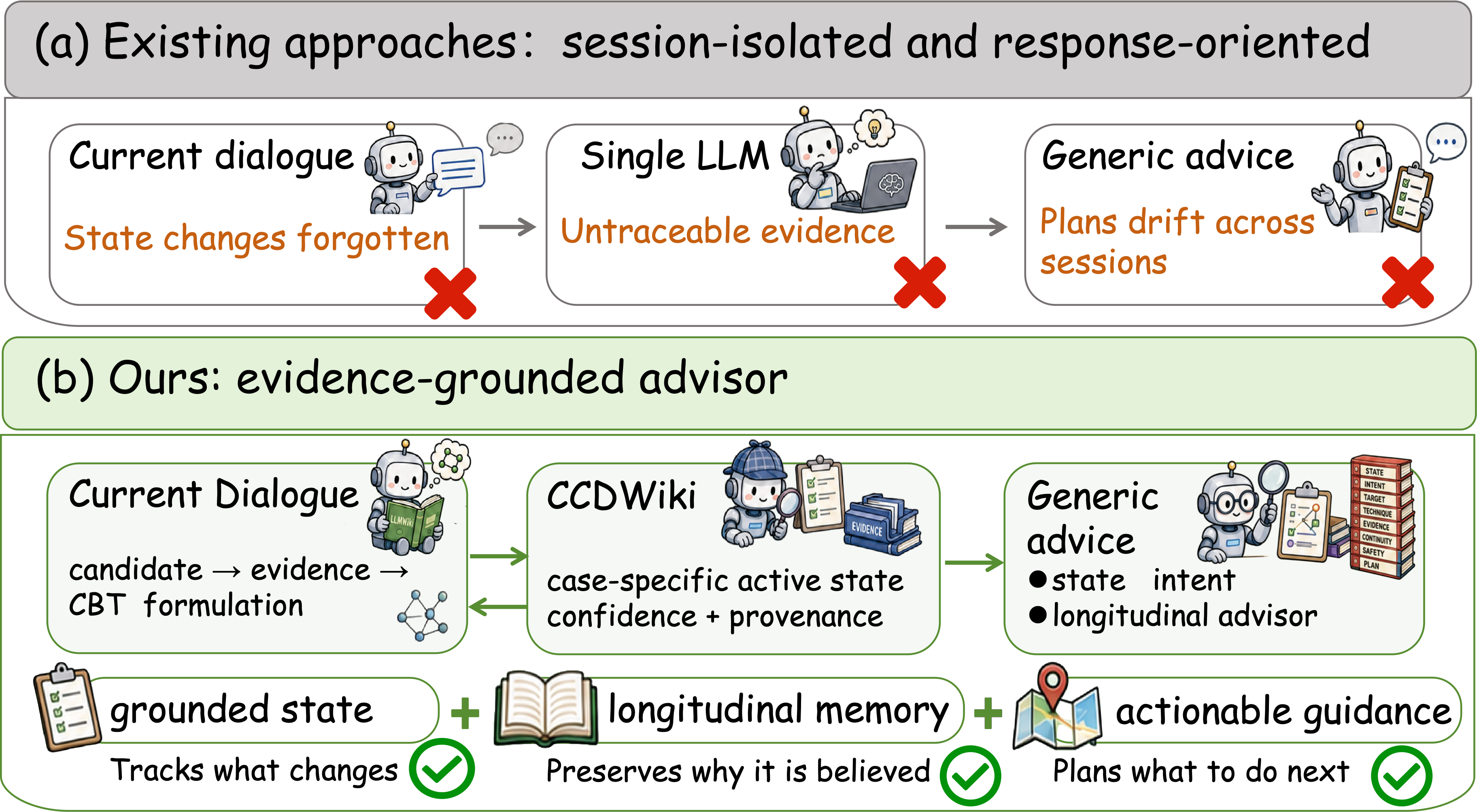}
\caption{Comparison between session-isolated response-oriented approaches and \textsc{CCDWiki}. Existing approaches can lose state changes, leave evidence untraceable, and yield intervention plans that drift across sessions. \textsc{CCDWiki} maintains a case-specific, evidence-grounded state with confidence and provenance, supporting longitudinal, actionable guidance.}
\label{fig:intro}
\end{figure}

Existing dialogue systems lack the architectural mechanisms required for evidential adjudication. Retrieval-augmented models recall past utterances to maintain conversational coherence \citep{lewis2020rag,packer2023memgpt}, while recent counseling-specific systems encode static client profiles or high-level strategy states \citep{lee2026psystep,cheng2022lookahead,zhang2025decoupledesc}. However, these systems treat dialogue history as a monolithic and unstructured context. They cannot distinguish between a \emph{tentative clinical guess} and an \emph{established fact}, nor can they trace the specific evidence supporting a given hypothesis. Figure~\ref{fig:intro} visualizes this gap: a session-isolated, response-oriented pipeline can forget state changes and leave evidence untraceable, causing intervention plans to drift across sessions. In CBT, acting on an unverified or contradicted hypothesis is not merely a generation error. It can damage the therapeutic alliance and lead to clinically harmful interventions.

To address this limitation, we propose \textsc{CCDWiki}, a plug-and-play advisor for longitudinal case conceptualization. It operates alongside a therapeutic LLM and requires no modification or retraining of the response backbone. Specifically, \textsc{CCDWiki} maintains a set of case-formulation hypotheses grounded in evidence from the client’s disclosures. Each new disclosure is evaluated against the current hypotheses. Supporting evidence strengthens a hypothesis, whereas insufficient evidence leaves it provisional. Contradictory evidence, in turn, prompts revision or removal. The formulation is therefore updated incrementally as the dialogue unfolds. At each turn, \textsc{CCDWiki} converts the current formulation into concise, case-specific guidance. This guidance is then provided to the therapeutic LLM, which generates the final response. In this way, case conceptualization is handled by an independent and revisable component, while response generation remains with the original backbone.

Our contributions are threefold:
\begin{enumerate}
\item We formulate longitudinal CBT case conceptualization as a process of evidence-grounded hypothesis adjudication.
\item We introduce \textsc{CCDWiki}, which organizes CCD hypotheses as evidence-linked, revisable entries and preserves their evolution across dialogue turns.
\item We develop a backbone-decoupled interface that supplies therapeutic LLMs with the current adjudicated formulation while leaving response generation to the original model.
\end{enumerate}

\section{Related Work}
\label{sec:related-work}

\textbf{Empathetic and Emotional Support Dialogue.}
EmpatheticDialogues established a benchmark for affect-sensitive open-domain
conversation \citep{rashkin2019empatheticdialogues}, and MoEL generates such
responses through a mixture of empathetic listeners \citep{lin2019moel}.
ESConv introduced emotional-support dialogues with explicit support strategies
\citep{liu2021esconv}. Lookahead planning then selects strategies by considering
their multi-turn consequences \citep{cheng2022lookahead}. CauESC incorporates
emotion causes and effects \citep{chen2024cauesc}. MAGO optimizes global
strategy sequences \citep{xie2025mago}. These methods improve how a system
chooses and realizes supportive moves. CCDWiki instead maintains the
longitudinal formulation that licenses those moves, including whether a
cognitive interpretation is provisional, supported, or revised.

\textbf{CBT-specific Dialogue Systems.}
CBT-LLM develops a CBT-oriented language model for mental-health question
answering \citep{na2024cbtllm}, while LLM4CBT aligns an LLM for CBT dialogue
\citep{kim2025llm4cbt}. DiaCBT provides a long-periodic counseling corpus
guided by the Cognitive Conceptualization Diagram (CCD)
\citep{zhou2025diacbt}. PSY-STEP separates surface problems from automatic
thoughts and represents the associated counseling action sequence
\citep{lee2026psystep}. TRACE uses two-stage reinforcement learning to explore
a causal graph before selecting a deeper psychological intervention
\citep{tang2026trace}. DMT-CBT maintains longitudinal therapeutic state,
including cognitive beliefs, emotional trajectories, and intervention progress
\citep{liu2026dmtcbt}. DecoupledESC separates strategy optimization from
response generation \citep{zhang2025decoupledesc}, and MentalGuide combines
state-aware planning with expert strategy distributions
\citep{he2026mentalguide}. Finally, multi-session counseling work models
therapy as an interaction that evolves across sessions
\citep{wang2026multisession}. These systems provide CBT data, aligned
generators, structured states, and policy mechanisms. CCDWiki isolates a
finer-grained object of update: every open-text formulation hypothesis retains
the client evidence that supports it, leaves it provisional, or overturns it.

\textbf{Modular Plugins for LLMs.}
PPDPP adds a trainable policy planner to a frozen LLM and selects the next
dialogue action \citep{deng2024ppdpp}. PsyAdvisor applies this plug-in pattern
to psychological conversations by prompting the original model with strategy
advice and proactive-questioning guidance \citep{hu2025psyadvisor}. LD-Agent
maintains event memory, a user persona, and an agent persona
\citep{li2025helloagain}. SeCom builds and retrieves compressed, semantically
coherent memory segments \citep{pan2025secom}. RMM creates multi-granularity
memories and refines retrieval with citation feedback \citep{tan2025rmm}.
MemORAI adds a provenance-enriched memory graph with query-adaptive retrieval
\citep{van2026memorai}. These systems either plan the next move or manage
information for response generation. 
\begin{figure*}[t]
\centering
\includegraphics[width=\textwidth]{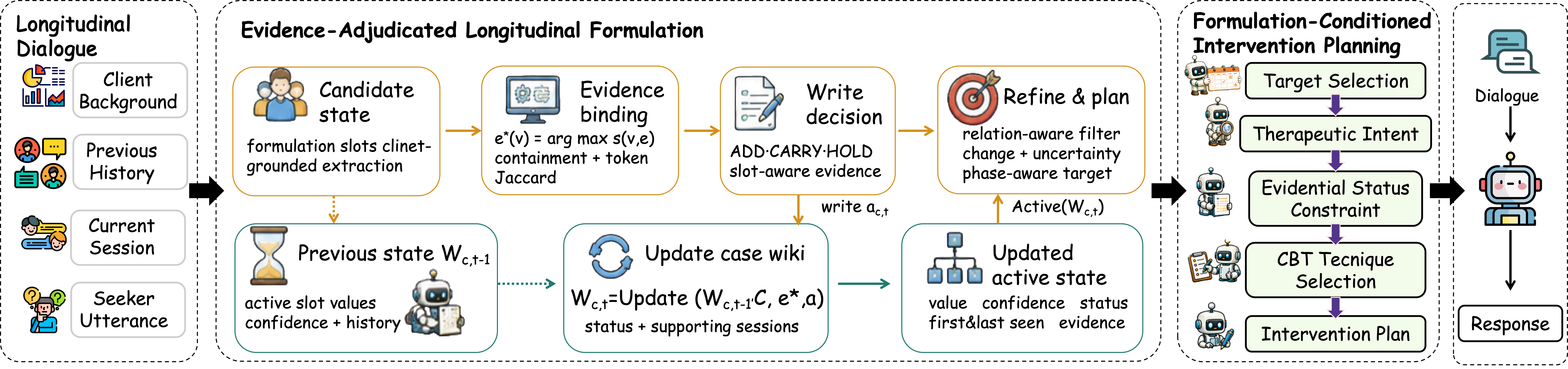}
\caption{Overview of \textsc{CCDWiki}. At response, client-attributable
evidence from the longitudinal dialogue updates a dynamic Cognitive
Conceptualization Diagram through Evidence-Adjudicated Longitudinal Formulation
(EALF). Formulation-Conditioned Intervention Planning (FCIP) then jointly selects
a formulation target, therapeutic intent, compatible CBT technique, and
case-specific guidance.}
\label{fig:overview}
\end{figure*}

\section{Method}
\label{sec:method}

\subsection{Overview}
\label{sec:method-overview}

As shown in Figure~\ref{fig:overview}, \textsc{CCDWiki} is a plug-and-play advisor that maintains an evidence-grounded Cognitive Conceptualization Diagram (CCD) and derives therapeutic decisions from it.
It consists of Evidence-Adjudicated Longitudinal Formulation (EALF) and Formulation-Conditioned Intervention Planning (FCIP).

\subsection{Problem Formulation}
\label{sec:problem-formulation}

At response point $k$, the observable information $\mathcal{O}_k$ consists of the client background, all preceding sessions, and the current session up to the latest client utterance.
The longitudinal case state is $\mathcal{S}_k = (\mathcal{F}_k, \mathcal{L}_k)$, where $\mathcal{F}_k$ is the live formulation and $\mathcal{L}_k$ its evidential lineage.
The generation of the updated state, an intervention plan $\mathbf{z}_k$, and the response $y_k$ is factorized as
\begin{equation}
\begin{aligned}
p(\mathcal{S}_k, \mathbf{z}_k, y_k \mid \mathcal{O}_k, \mathcal{S}_{k-1})
&= p(\mathcal{S}_k \mid \mathcal{O}_k, \mathcal{S}_{k-1}) \\
&\quad \times p(\mathbf{z}_k \mid \mathcal{O}_k, \mathcal{S}_k) \\
&\quad \times p_{\theta}(y_k \mid d_k, \mathbf{z}_k),
\end{aligned}
\label{eq:task-factorization}
\end{equation}
where $d_k$ denotes the dialogue prefix ending with the triggering client turn.
\textsc{CCDWiki} instantiates the first two factors as explicit reasoning modules and keeps the last factor as a frozen therapeutic language model.

\subsection{Evidence-Adjudicated Longitudinal Formulation}
\label{sec:ealf}

A longitudinal formulation is represented as an evolving system of hypotheses rather than a fixed summary of client attributes. Each hypothesis expresses a possible element or relation within the CCD and remains distinguishable from the evidence on which it is based. This separation allows several disclosures to jointly support an interpretation, while also allowing later evidence to qualify or displace an interpretation established earlier.

At response point $k$, the live formulation is
\begin{equation}
\mathcal{F}_{k}
=
\left(
\mathcal{P}_{k},
\mathcal{A}_{k}
\right),
\label{eq:formulation-state}
\end{equation}
where $\mathcal{P}_{k}$ contains provisional hypotheses and $\mathcal{A}_{k}$ contains hypotheses admitted into the current working formulation. A formulation hypothesis is represented as
\begin{equation}
h_i
=
\left(
q_i,\,
v_i,\,
W_i,\,
\pi_i
\right),
\label{eq:formulation-hypothesis}
\end{equation}
where $q_i$ denotes its CCD dimension, $v_i$ is its propositional content, $W_i$ is its source-resolved evidence record, and $\pi_i$ identifies its predecessor when the hypothesis results from a revision. The CCD dimensions cover clinically relevant situations, automatic thoughts, emotions, behaviours, beliefs, coping patterns, maintaining relations, and protective factors.

Let $\mathcal{E}_{k}^{\mathrm{new}}$ denote the client-attributable evidence introduced at response point $k$. Each evidence item comprises an observed text span and its source coordinate:
\begin{equation}
e
=
\left(
\operatorname{span}(e),\,
c,\,
t(e),\,
j(e)
\right).
\label{eq:evidence-item}
\end{equation}
Therapist utterances may contextualize a disclosure but do not themselves constitute evidence for a client-specific formulation claim.

Candidate hypotheses are induced from the new evidence in relation to the preceding formulation:
\begin{equation}
\mathcal{C}_{k}
=
\operatorname{Induce}
\left(
\mathcal{E}_{k}^{\mathrm{new}},
\mathcal{F}_{k-1}
\right).
\label{eq:candidate-induction}
\end{equation}
A candidate may instantiate a previously unrepresented CCD element, provide an alternative interpretation of an existing element, or refine the content of an earlier hypothesis. Candidates that express the same interpretation within the same CCD dimension are aligned with the corresponding hypothesis; materially different interpretations remain separate, while qualifications of an existing interpretation are linked to it as potential successors.

For every hypothesis--evidence pair, EALF determines how the observed evidence bears on the hypothesis:
\begin{equation}
r(h_i,e)
\in
\left\{
\textsc{Support},
\textsc{Qualify},
\textsc{Contradict},
\textsc{Irrelevant}
\right\}.
\label{eq:evidence-relation}
\end{equation}
Accordingly, the evidence record of $h_i$ is partitioned as
\begin{equation}
W_i^{k}
=
\left(
W_{i,+}^{k},
W_{i,\circ}^{k},
W_{i,-}^{k}
\right),
\label{eq:evidence-record}
\end{equation}
where $W_{i,+}^{k}$ contains supporting evidence, $W_{i,\circ}^{k}$ contains qualifying evidence, and $W_{i,-}^{k}$ contains contradictory evidence. Irrelevant observations do not alter the evidential standing of the hypothesis.

The transition from a plausible interpretation to an active formulation hypothesis is governed by its accumulated evidential basis:
\begin{equation}
h_i\in\mathcal{A}_{k}
\quad\Longleftrightarrow\quad
\rho
\left(
h_i,\,
W_{i,+}^{k},\,
W_{i,\circ}^{k},\,
W_{i,-}^{k}
\right)
\geq
\tau_{q_i},
\label{eq:hypothesis-admission}
\end{equation}
where $\rho$ evaluates whether the available evidence coherently supports the content of $h_i$, and $\tau_{q_i}$ is the admission criterion associated with its CCD dimension. Evidence distributed across different turns or sessions may therefore converge on the same hypothesis, whereas an interpretation that remains plausible but insufficiently substantiated stays in $\mathcal{P}_{k}$.

The evidential record continues to regulate a hypothesis after its admission. For each live hypothesis, EALF assigns one of five lifecycle outcomes:
\begin{equation}
\delta_k(h_i)
\in
\left\{
\textsc{Provisional},
\textsc{Commit},
\textsc{Carry},
\textsc{Revise},
\textsc{Retire}
\right\}.
\label{eq:lifecycle-outcomes}
\end{equation}
\textsc{Provisional} retains an insufficiently supported hypothesis in $\mathcal{P}_{k}$; \textsc{Commit} admits a hypothesis into $\mathcal{A}_{k}$; and \textsc{Carry} preserves an active hypothesis whose interpretation remains compatible with the accumulated evidence. \textsc{Revise} replaces a live hypothesis with a supported successor, while \textsc{Retire} removes an interpretation whose evidential basis no longer warrants its use in the working formulation.

A revision is expressed as
\begin{equation}
\textsc{Revise}
\left(
h_i\rightarrow h_i'
\right),
\qquad
\pi(h_i')=h_i.
\label{eq:hypothesis-revision}
\end{equation}
The successor $h_i'$ remains within the same CCD dimension but expresses the interpretation warranted by the later evidence. Evidence attached to the predecessor remains accessible through $\mathcal{L}_{k}$ and is not automatically reassigned as support for the successor. A proposed successor replaces its predecessor only when evidence bearing on its own content satisfies the corresponding admission criterion; otherwise, it remains provisional while the predecessor retains its current standing.

Let $\mathcal{B}_{k}$ denote the set of adjudicated lifecycle decisions at response point $k$. The longitudinal state evolves as
\begin{equation}
\left(
\mathcal{F}_{k},
\mathcal{L}_{k}
\right)
=
T
\left(
\mathcal{F}_{k-1},
\mathcal{L}_{k-1},
\mathcal{B}_{k}
\right),
\label{eq:formulation-transition}
\end{equation}
where $T$ applies the admitted changes to the live formulation and appends their evidential justifications to the lineage. Revised and retired hypotheses consequently leave the active formulation without being erased from its reasoning history.

Through this process, EALF maintains a formulation that is stable enough to support intervention planning while remaining revisable under subsequent evidence. The live formulation records which interpretations are currently warranted; the lineage records how that warrant was established and how it has changed over time.

\subsection{Formulation-Conditioned Intervention Planning}
\label{sec:fcip}

The formulation established by EALF defines the system's current understanding of the case, but does not uniquely determine the next therapeutic action. FCIP converts that understanding into an intervention decision by jointly selecting a formulation target, a therapeutic intent, and a compatible CBT technique.

An intervention plan is represented as
\begin{equation}
\mathbf{z}_{k}
=
\left(
h_k^{*},\,
m_k^{*},\,
a_k^{*},\,
g_k
\right),
\label{eq:intervention-plan}
\end{equation}
where $h_k^{*}$ is the targeted formulation element or relation, $m_k^{*}$ is the immediate therapeutic intent, $a_k^{*}$ is the selected CBT technique, and $g_k$ is the case-specific guidance supplied to the response model.

Target selection operates over the relational structure of the current CCD. The planner considers not only individual situations, thoughts, emotions, beliefs, and behaviours, but also the relations through which these elements contribute to the client's presenting difficulty. A target is therefore selected according to its relevance to the latest disclosure, its position within the working formulation, and its therapeutic importance at the current stage of the interaction.

The therapeutic intent space is defined as
\begin{equation}
\mathcal{M}
=
\left\{
\textsc{Explore},
\textsc{Test},
\textsc{Restructure}
\right\}.
\label{eq:therapeutic-intents}
\end{equation}
\textsc{Explore} seeks information required to clarify or extend the formulation; \textsc{Test} examines the validity, scope, or consequences of an admitted interpretation; and \textsc{Restructure} intervenes on a supported cognitive or behavioural pattern.

The admissible intent depends on the evidential standing of the target:
\begin{equation}
\Gamma_k(h)
=
\begin{cases}
\left\{
\textsc{Explore}
\right\},
&
h\in\mathcal{P}_{k},
\\[4pt]
\left\{
\textsc{Explore},
\textsc{Test},
\textsc{Restructure}
\right\},
&
h\in\mathcal{A}_{k}.
\end{cases}
\label{eq:intent-constraint}
\end{equation}
A provisional hypothesis may guide further inquiry, but it is not treated as an established premise for hypothesis testing or restructuring. Active hypotheses may support stronger therapeutic actions, subject to their CCD role, evidential record, and the current phase of the dialogue.

Let $\Omega(m)$ denote the family of CBT techniques compatible with intent $m$. Exploration may be realized through clarification, guided discovery, or focused evidence elicitation; testing may employ Socratic questioning, evidence examination, or behavioural experiments; and restructuring may employ cognitive reappraisal, alternative-thought construction, behavioural activation, or modification of a supported maintaining process. The admissible planning space is
\begin{equation}
\mathcal{Z}_{k}
=
\left\{
(h,m,a):
h\in\mathcal{P}_{k}\cup\mathcal{A}_{k},\,
m\in\Gamma_k(h),\,
a\in\Omega(m)
\right\}.
\label{eq:admissible-plans}
\end{equation}

The target, intent, and technique are selected jointly:
\begin{equation}
\left(
h_k^{*},
m_k^{*},
a_k^{*}
\right)
=
\operatorname*{arg\,max}_{(h,m,a)\in\mathcal{Z}_{k}}
S
\left(
h,m,a
\mid
\mathcal{O}_{k},
\mathcal{F}_{k},
\mathcal{L}_{k}
\right),
\label{eq:intervention-selection}
\end{equation}
where $S$ evaluates the relevance of the target to the current disclosure, its role within the CCD, the evidential basis available for acting on it, the compatibility between therapeutic intent and technique, and the continuity of the proposed intervention with preceding therapeutic work.

The selected decision is subsequently expressed as case-specific response guidance:
\begin{equation}
g_k
=
R
\left(
h_k^{*},
W_k(h_k^{*}),
m_k^{*},
a_k^{*},
\mathcal{F}_{k}
\right).
\label{eq:guidance-construction}
\end{equation}
The guidance identifies the formulation content to be addressed, the evidence relevant to that content, the immediate therapeutic objective, the selected CBT technique, and the intended direction of the response. It therefore specifies a therapeutic decision rather than providing a generic summary of the client or dialogue history.

Finally, the frozen therapeutic language model realizes the intervention plan as
\begin{equation}
y_k
\sim
p_{\theta}
\left(
y
\mid
d_{c,t(k),\leq j(k)},
\mathbf{z}_{k}
\right),
\qquad
\theta\ \mathrm{fixed}.
\label{eq:response-realization}
\end{equation}
The formulation layer determines which case interpretations are currently licensed by the evidence, and the planning layer determines how those interpretations should govern the next therapeutic move. The response model is thereby responsible for expressing an already specified intervention, rather than independently reconstructing the longitudinal case formulation and therapeutic decision from the complete dialogue history.

\begin{table*}[t]
\centering
\footnotesize
\setlength{\tabcolsep}{3pt}
\begin{tabular}{llc|ccccccc}
\toprule
Type & Models & \mbox{\textsc{CCDWiki}} & Understand. & Interpers. & Collabor. & Guided disc. & Focus & Strategy & Mean \\
\midrule
\multirow{4}{*}{\begin{tabular}{@{}c@{}}Large-scale\\General-purpose\\LLMs\end{tabular}} 
 & \multirow{2}{*}{DeepSeek V4 Pro} & \(\times\) & 3.81 & 4.19 & 3.08 & 3.16 & 4.00 & 3.44 & 3.62 \\
 & & \(\checkmark\) & 3.86 & 4.01 & 2.48 & 3.51 & 4.23 & 3.65 & 3.62 \\
 & \multirow{2}{*}{GPT-5.4} & \(\times\) & 3.99 & 4.11 & 3.42 & 3.82 & 3.95 & 3.41 & 3.78 \\
 & & \(\checkmark\) & 3.93 & 4.00 & 3.46 & 3.86 & 4.00 & 3.75 & 3.83 \\
\midrule
\multirow{6}{*}{\begin{tabular}{@{}c@{}}Small-scale\\Open-source\\LLMs\end{tabular}} 
 & \multirow{2}{*}{GLM4 9B} & \(\times\) & 2.35 & 3.96 & 2.89 & 2.10 & 1.76 & 2.48 & 2.59 \\
 & & \(\checkmark\) & 3.43 & 3.99 & 3.22 & 2.92 & 3.77 & 3.24 & 3.43 \\
 & \multirow{2}{*}{InternLM3 8B} & \(\times\) & 2.91 & 3.92 & 3.10 & 2.22 & 2.65 & 2.70 & 2.91 \\
 & & \(\checkmark\) & 3.60 & 4.02 & 3.42 & 2.58 & 3.78 & 3.18 & 3.43 \\
 & \multirow{2}{*}{Qwen3 8B} & \(\times\) & 3.39 & 3.87 & 2.1 & 3.13 & 3.46 & 2.55 & 3.10 \\
 & & \(\checkmark\) & 3.78 & 3.96 & 1.87 & 3.69 & 3.73 & 2.60 & 3.27 \\
\midrule
\multirow{6}{*}{\begin{tabular}{@{}c@{}}Psychological\\LLMs\end{tabular}} 
 & \multirow{2}{*}{SoulChat} & \(\times\) & 3.65 & 4.10 & 3.86 & 3.15 & 2.95 & 3.02 & 3.46 \\
 & & \(\checkmark\) & 3.78 & 4.05 & 3.72 & 3.55 & 3.38 & 3.47 & 3.66 \\
 & \multirow{2}{*}{CBT-LLM} & \(\times\) & 3.72 & 3.65 & 3.35 & 3.52 & 3.86 & 3.78 & 3.65 \\
 & & \(\checkmark\) & 3.84 & 3.67 & 3.41 & 3.76 & 4.12 & 4.03 & 3.81 \\
 & \multirow{2}{*}{PsyChat} & \(\times\) & 3.55 & 4.08 & 3.83 & 3.20 & 3.00 & 3.12 & 3.46 \\
 & & \(\checkmark\) & 3.68 & 4.02 & 3.72 & 3.56 & 3.36 & 3.48 & 3.64 \\
\bottomrule
\end{tabular}
\caption{CTRS profiles for eight backbones with and without \mbox{\textsc{CCDWiki}}.}
\label{tab:ctrs-results}
\end{table*}

\section{Experiments}
\label{sec:experiments}

\subsection{Experimental Setup}

\paragraph{Evaluation scope.}
We evaluate CCDWiki as a plug-and-play guidance method for longitudinal CBT
dialogue generation. CTRS and PANAS results are available for eight Therapist
backbones. DeepSeek V4 Pro and GPT-5.4 form the large-scale general-purpose
group. GLM4-9B, InternLM3-8B, and Qwen3-8B form the small-scale open-source
group. SoulChat, CBT-LLM, and PsyChat form the psychological LLM group. UED summaries are
available for the five general-purpose and small-scale backbones. For every
backbone, we compare a Base condition with a CCDWiki-guided condition.

The analysis focuses on response-side CBT competence and simulated affective
trajectories. CTRS, UED, and PANAS do not directly measure whether an internal
formulation hypothesis was correctly admitted, carried, revised, or retired,
and none of the reported scores constitutes a clinical outcome.

\paragraph{Dataset and split.}
We use the CBT subset of PsychEval \citep{pan2026psycheval}, which contains 148
simulated Chinese counselling cases, 1,112 sessions, and 50,018 dialogue turns.
Each case contains five to ten sessions. We construct one longitudinal instance
for every non-final session. Each instance pairs the observed dialogue with
silver fields for the current formulation, subsequent change, and next-session
target. Future-session content
is used only to construct these reference fields and is excluded from
generation-time observations.

The split yields 964 longitudinal instances. The
dialogue-generation comparison uses the same 22 held-out evaluation cases for
all backbones and both conditions. The instance counts document Advisor
supervision and corpus construction; they are not denominators for the aggregate
response scores below.

\paragraph{Dialogue-generation protocol.}
Each evaluation case contains 35 turns. Turns 1--10 are frozen
Client--Therapist pairs retained in their original speaking order. Turns
11--35 are generated sequentially. At generated turn $t$, the visible history
contains at most 12 turns:
\begin{equation}
H_t = \{d_{\max(1,t-12)},\ldots,d_{t-1}\},
\end{equation}
Here, $d_i=(s_i,c_i,r_i)$ contains the system instruction, Client utterance,
and Therapist response. Thus the first generated turn uses all ten frozen
turns, and subsequent turns use at most the most recent 12 complete turns.

Grok~4.5 generates the current Client utterance while preserving case identity,
known facts, symptoms, and the observed pace of change. The Client is instructed
to respond to the latest Therapist turn and introduce a new reaction, meaning,
difficulty, or intervention outcome rather than paraphrase a recent utterance.
In Base, the Therapist's system field is empty and no Advisor is called. Under
CCDWiki, a Qwen3.5-4B Advisor reads the visible history and current Client turn
and generates turn-specific Guidance, which is combined with a concise CBT role
instruction. The selected Therapist then receives $H_t$, the current Client
utterance, and the condition-specific system instruction.


\paragraph{Evaluation measures.}
The Cognitive Therapy Rating Scale (CTRS) covers understanding, interpersonal
effectiveness, collaboration, guided discovery, CBT focus, and CBT strategy
\citep{affrunti2019ctrsfactor}. We report the six dimensions and their
unweighted mean, with higher values indicating stronger CBT competence under
the scorer. UED summarizes the mean and variability of matched emotion signals
together with Valence, Arousal, and Dominance. These dimensions characterize
trajectory shape and have no common favorable direction. PANAS contains ten
Positive Affect and ten Negative Affect items \citep{watson1988panas}. We
report their separate unweighted means of scorer-estimated after--before
changes rather than collapsing the two polarities into one score.

\paragraph{Implementation details.}
The CCDWiki Advisor uses Qwen3.5-4B as its backbone. Dynamic supervision is
constructed with GPT-5.4-mini from the case-disjoint training and development
partitions. Each example contains up to 12 preceding turns, the current Client
utterance, and the visible CCDWiki state. The source Therapist response is
available only to the teacher when constructing the target Guidance and is
excluded from Advisor inference. We train the Advisor for three epochs using
4-bit NF4 QLoRA. The LoRA rank is 16, the scaling factor is 32, and dropout is
0.05. Optimization uses a learning rate of $1\times10^{-4}$, per-device batch
size of 1, and gradient accumulation over eight steps. We set the maximum
sequence length to 1,792 and use bfloat16 computation with gradient
checkpointing. Adapters are applied to the query, key, value, and output
projection modules. At inference, the Advisor uses temperature 0.2 and a
maximum output budget of 180 tokens. Grok~4.5 uses temperature 0.65 and a
maximum budget of 256 tokens. Therapist models use temperature 0.45, a maximum
budget of 256 tokens, and a repetition penalty of 1.08 where supported.

\subsection{CTRS Results}

Table~\ref{tab:ctrs-results} reports CBT competence for all eight backbones.
The CCDWiki condition has an equal or higher six-dimension mean at the displayed
precision for every backbone, although the size and dimension-level consistency
of the change vary.

At the displayed precision, the largest mean increases occur for GLM4-9B
($+0.84$) and InternLM3-8B ($+0.52$). They are followed by SoulChat ($+0.20$),
PsyChat ($+0.18$), Qwen3-8B ($+0.17$), CBT-LLM ($+0.16$), and GPT-5.4
($+0.05$).
DeepSeek is unchanged at two-decimal precision. The gains for GLM4-9B and
InternLM3-8B span all six dimensions and are largest in CBT focus ($+2.01$ and
$+1.13$, respectively). CBT-LLM also improves on all six dimensions, indicating
that CCDWiki remains complementary to a backbone already specialized for CBT.

The aggregate means conceal several trade-offs. DeepSeek decreases in
interpersonal effectiveness ($-0.18$) and collaboration ($-0.60$), GPT-5.4
decreases in understanding ($-0.06$) and interpersonal effectiveness
($-0.11$), and Qwen3-8B decreases in collaboration ($-0.23$). SoulChat and
PsyChat also trade higher guided discovery, focus, and strategy for lower
interpersonal effectiveness and collaboration. A positive aggregate contrast
therefore does not imply uniform improvement across CBT competencies.

\subsection{UED Results}

\begin{figure}[t]
\centering
\includegraphics[width=\columnwidth]{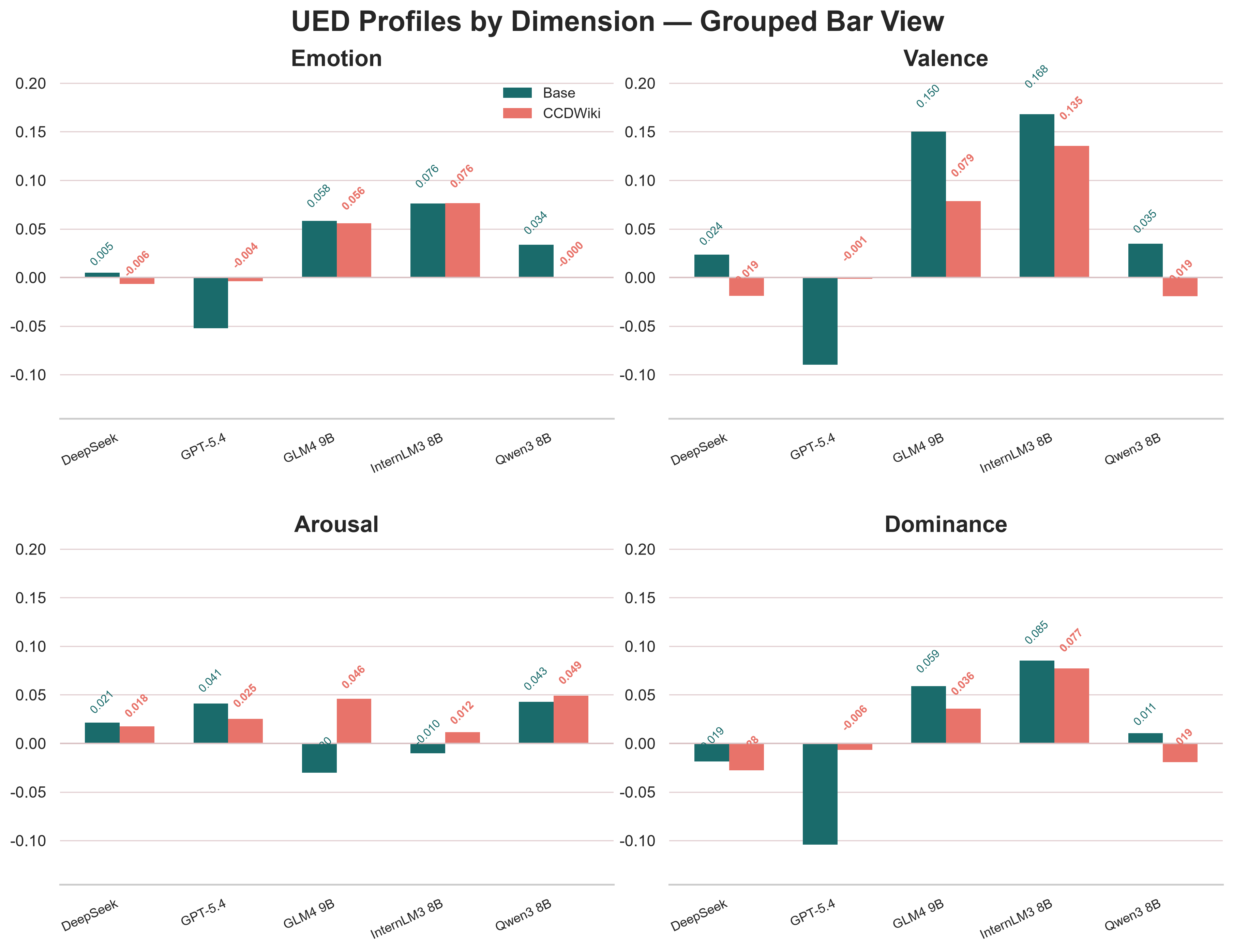}
\caption{Normalized UED profiles for four Therapist backbones under Base and
CCDWiki. Each spoke is min--max normalized over the eight displayed
model--condition profiles. Radial positions support comparisons on the same
spoke but are not raw UED values, and greater radial distance does not imply
better response quality because the descriptors have no common favorable
direction. Panels show (a) DeepSeek V4 Pro, (b) GLM4-9B,
(c) InternLM3-8B, and (d) Qwen3-8B.}
\label{fig:ued-results}
\end{figure}

Figure~\ref{fig:ued-results} visualizes the corresponding Base--CCDWiki profile
changes for the four non-GPT backbones. UED results are not available for the
three psychological LLMs and are therefore not inferred from their CTRS or
PANAS scores.

GLM4-9B changes most strongly in Arousal ($+0.0820$) and Variability
($+0.0053$), while its Valence decreases by $0.0708$. InternLM3-8B retains an
almost unchanged Emotion mean ($-0.0007$) while Variability and Arousal
increase by $0.0016$ and $0.0280$, respectively. DeepSeek changes little in
Arousal and Dominance, but its Emotion mean and Valence decrease by $0.0160$
and $0.0430$. Qwen3-8B likewise shows lower Emotion mean ($-0.0257$), Valence
($-0.0499$), and Dominance ($-0.0313$), alongside small increases in
Variability and Arousal.

Valid-row counts differ for DeepSeek, Qwen3-8B, and GPT-5.4. GPT-5.4 covers 56
rows from 20 Base cases and 62 rows from 22 CCDWiki cases; DeepSeek and
Qwen3-8B also have unequal row counts across conditions. These all-available
summaries are not paired case-level contrasts, so their differences may partly
reflect which rows were retained. We therefore use UED to describe trajectory
shape rather than for strict cross-model ranking or isolated improvement
claims.

\subsection{PANAS Results}

Table~\ref{tab:panas-results} reports scorer-estimated Positive and Negative
Affect changes for all eight backbones. Positive Affect increases and Negative
Affect decreases are favorable directions, but the two polarities are analyzed
separately.

\begin{table}[t]
\centering
\footnotesize
\setlength{\tabcolsep}{4.5pt}
\renewcommand{\arraystretch}{1.05}

\begin{tabular}{llcc}
\toprule
Model & Setting & Positive mean & Negative mean \\
\midrule

\multirow{2}{*}{DeepSeek V4 Pro}
    & Base    & 0.44 & 0.41 \\
    & CCDWiki & 0.58 & 0.39 \\
\cmidrule(lr){2-4}

\multirow{2}{*}{GPT-5.4}
    & Base    & 0.19 & 0.51 \\
    & CCDWiki & 0.12 & 0.53 \\

\midrule

\multirow{2}{*}{GLM4-9B}
    & Base    & 0.17 & 0.17 \\
    & CCDWiki & 0.63 & 0.33 \\
\cmidrule(lr){2-4}

\multirow{2}{*}{InternLM3-8B}
    & Base    & 0.31 & 0.39 \\
    & CCDWiki & 0.73 & 0.21 \\
\cmidrule(lr){2-4}

\multirow{2}{*}{Qwen3-8B}
    & Base    & 0.30 & 0.50 \\
    & CCDWiki & 0.52 & 0.52 \\

\midrule

\multirow{2}{*}{SoulChat}
    & Base    & 0.49 & 0.43 \\
    & CCDWiki & 0.60 & 0.38 \\
\cmidrule(lr){2-4}

\multirow{2}{*}{CBT-LLM}
    & Base    & 0.42 & 0.34 \\
    & CCDWiki & 0.65 & 0.25 \\
\cmidrule(lr){2-4}

\multirow{2}{*}{PsyChat}
    & Base    & 0.50 & 0.41 \\
    & CCDWiki & 0.62 & 0.35 \\

\bottomrule
\end{tabular}
\caption{PANAS after--before affect-change scores under Base and CCDWiki.
Each column reports the mean over ten polarity-specific items. The values
characterize simulated dialogues rather than client-reported clinical outcomes.}
\label{tab:panas-results}
\end{table}

At the displayed precision, CCDWiki changes Positive Affect by $+0.14$ for
DeepSeek, $+0.46$ for GLM4-9B, $+0.42$ for InternLM3-8B, and $+0.22$ for
Qwen3-8B. The corresponding changes are $+0.11$ for SoulChat, $+0.23$ for
CBT-LLM, $+0.12$ for PsyChat, and $-0.07$ for GPT-5.4. Negative Affect changes
by $-0.02$, $+0.16$, $-0.18$, $+0.02$, $-0.05$, $-0.09$, $-0.06$, and
$+0.02$, respectively.

InternLM3-8B has the clearest balanced numerical profile among the five
non-psychological backbones: higher Positive Affect coincides with the largest
decrease in Negative Affect. DeepSeek shows the same directions at smaller
magnitudes. GLM4-9B's largest Positive Affect increase coincides with higher
Negative Affect. Qwen3-8B combines a moderate Positive Affect increase with a
small Negative Affect increase. GPT-5.4 moves in the unfavorable direction on
both aggregates. All three psychological LLMs combine higher Positive Affect
with lower Negative Affect, with CBT-LLM showing the largest joint change in
this group.

\subsection{Cross-Metric Analysis}

Across the general-purpose and small-scale backbones, InternLM3-8B exhibits the
clearest balanced response-side gain. Its CTRS mean increases by $0.52$,
Positive Affect increases by $0.42$, and Negative Affect decreases by $0.18$.
Its UED Emotion mean remains nearly stable while Variability rises slightly.
GLM4-9B records the largest CTRS and Positive Affect increases, but these
coincide with higher UED Arousal and higher Negative Affect. DeepSeek retains a
comparatively strong absolute CTRS level but changes little in aggregate CBT
competence. Its PANAS profile moves modestly in both favorable directions,
whereas its UED Emotion mean and Valence are lower under CCDWiki.

Qwen3-8B shows moderate gains in the CTRS mean and Positive Affect, accompanied
by lower collaboration and a small increase in Negative Affect. GPT-5.4 retains
the highest absolute CTRS mean among the five non-psychological backbones, but
its small competence gain is not accompanied by an affective advantage over
Base. Among the psychological LLMs, all three backbones improve in aggregate
CTRS and both PANAS directions. CBT-LLM is the most consistent of the three. It
improves on every CTRS dimension and has the largest psychological-model
changes in both PANAS directions. SoulChat and PsyChat show favorable aggregate
changes but retain
dimension-level trade-offs in interpersonal effectiveness and collaboration.
These results indicate heterogeneous, backbone-dependent response profiles
rather than a universal improvement pattern. The comparison remains
descriptive because case-level paired scores and repeated stochastic rollouts
are unavailable. 

\begin{table}[t]
\centering
\scriptsize
\setlength{\tabcolsep}{2.2pt}
\begin{tabular}{l|ccccccc}
\toprule
Condition & Und. & Inter. & Collab. & Guide. & Focus & Strat. & Mean \\
\midrule
GPT-5.4 & 3.99 & 4.11 & 3.42 & 3.82 & 3.95 & 3.41 & 3.78 \\
w/o EALF & 3.95 & 4.04 & 3.44 & 3.77 & 3.91 & 3.57 & 3.78 \\
w/o FCIP & 3.95 & 4.02 & 3.45 & 3.82 & 3.96 & 3.54 & 3.79 \\
Full \mbox{\textsc{CCDWiki}} & 3.93 & 4.00 & 3.46 & 3.86 & 4.00 & 3.75 & 3.83 \\
\bottomrule
\end{tabular}
\caption{GPT component ablation on CTRS. EALF: Evidence-Adjudicated
Longitudinal Formulation; FCIP: Formulation-Conditioned Intervention Planning.}
\label{tab:ablation}
\end{table}

\subsection{Component Ablation}

Table~\ref{tab:ablation} is for internal review. Its base and full rows reuse
the paired GPT-5.4 values in Table~\ref{tab:ctrs-results}; the two ablated rows
are candidate values awaiting experimental confirmation and are not independently
verified effects. Relative to the full draft configuration, removing EALF lowers guided discovery
and CBT focus by $0.09$ each and CBT strategy by $0.18$. This pattern is
consistent with the intended role of an evidence-adjudicated formulation: it
keeps exploratory questions and focal interventions tied to a current,
longitudinally supported case hypothesis. Removing FCIP leaves the formulation
available but lowers CBT strategy by $0.21$, the largest component-level
contrast in the table. The proposed interpretation is that formulation context
alone does not fully specify a therapeutic move; explicit target--intent--technique
selection is needed to translate it into a strategy-bearing response.

\section{Conclusion}
\label{sec:conclusion}

Longitudinal CBT dialogue requires more than access to prior exchanges; it depends on an explicit, evolving case formulation in which hypotheses remain traceable to client evidence and can be strengthened, revised, or withdrawn as new information emerges. We therefore introduced \textsc{CCDWiki}, a backbone-agnostic advisor that represents case conceptualization as evidence-grounded hypothesis adjudication, preserves the lineage and status of formulation entries across sessions, and converts the current formulation into concise guidance for a therapeutic LLM without modifying its parameters or response mechanism. This separation makes longitudinal clinical reasoning more transparent and revisable while retaining the generative flexibility of the underlying model. 

\bibliography{aaai2027}


\end{document}